\documentclass[10pt,twocolumn,letterpaper]{article}

\usepackage{cvpr}
\usepackage{times}
\usepackage{epsfig}
\usepackage{graphicx}
\usepackage{graphics}
\usepackage{amsmath}
\usepackage{amssymb}
\usepackage{subcaption}
\usepackage[breaklinks=true,bookmarks=false]{hyperref}

\cvprfinalcopy % *** Uncomment this line for the final submission

\def\cvprPaperID{00} % *** Enter the CVPR Paper ID here

\begin{document}

%%%%%%%%% TITLE
\title{Self-Transfer Learning for Fully Weakly Supervised Object Localization}
%\title{Self-Transfer Learning: Building-Up Localization Seed from Classification Network for Authentic Weakly-Supervised Localization}

\author{
\vspace{3mm}
Sangheum Hwang and Hyo-Eun Kim\\
Lunit Inc., Seoul, Korea\\
{\tt\small \{shwang, hekim\}@lunit.io}
%\and
%Hyo-Eun Kim\\
%Lunit Inc., Seoul, Korea\\
%{\tt\small hekim@lunit.io}
}

\maketitle
%\thispagestyle{empty}

%%%%%%%%% ABSTRACT
\begin{abstract}
Recent advances of deep learning have achieved remarkable performances in various challenging computer vision tasks. Especially in object localization, deep convolutional neural networks outperform traditional approaches based on extraction of data/task-driven features instead of hand-crafted features. Although location information of region-of-interests (ROIs) gives good prior for object localization, it requires heavy annotation efforts from human resources. Thus a weakly supervised framework for object localization is introduced. The term ``weakly'' means that this framework only uses image-level labeled datasets to train a network. With the help of transfer learning which adopts weight parameters of a pre-trained network, the weakly supervised learning framework for object localization performs well because the pre-trained network already has well-trained class-specific features. However, those approaches cannot be used for some applications which do not have pre-trained networks or well-localized large scale images. Medical image analysis is a representative among those applications because it is impossible to obtain such pre-trained networks. In this work, we present a ``fully'' weakly supervised framework for object localization (``semi''-weakly is the counterpart which uses pre-trained filters for weakly supervised localization) named as self-transfer learning (STL). It jointly optimizes both classification and localization networks simultaneously. By controlling a supervision level of the localization network, STL helps the localization network focus on correct ROIs without any types of priors. We evaluate the proposed STL framework using two medical image datasets, chest X-rays and mammograms, and achieve signiticantly better localization performance compared to previous weakly supervised approaches.
\end{abstract}

\begin{figure*}
\begin{center}
\includegraphics[width=\textwidth]{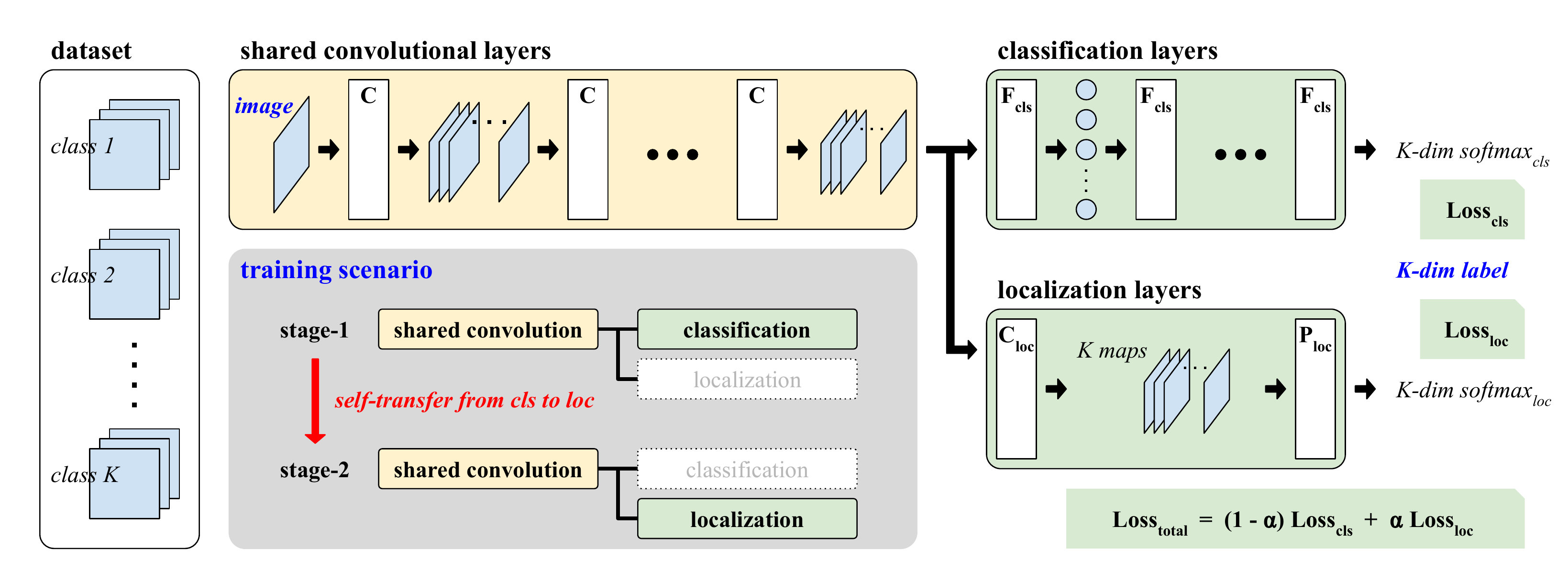}
\end{center}
   \caption{Overall architecture and training scenario (\textbf{C} in shared convolutional layers: convolutional or convolutional with a max pooling, \textbf{F}$_{\textbf{cls}}$ in classification layers: fully connected or fully connected with a dropout regularization, \textbf{C}$_{\textbf{loc}}$ and \textbf{P}$_{\textbf{loc}}$ in localization layers: 1$\times$1 convolutional layer and global pooling layer respectively).  \textbf{Loss}$_{\textbf{cls}}$ and \textbf{Loss}$_{\textbf{loc}}$ are cross-entropy losses between the true label and \textit{softmax} outputs from both classification and localization layers, and the final objective function \textbf{Loss}$_{\textbf{total}}$ is an weighted sum of those losses with a controllable hyperparameter $\mathbf{\alpha}$. Self-transfer learning is realized by re-weighting the $\mathbf{\alpha}$ adaptively in a training phase.}
\label{fig:overall_arch}
\end{figure*}

%%%%%%%%% BODY TEXT
\section{Introduction}
Recently, deep convolutional neural networks (CNN) show promising performances in various computer vision tasks such as object classification~\cite{Krizhevsky2012Classification,Szegedy2015Classification}, localization (or detection)~\cite{Sermanet2013Detection,Erhan2014Detection}, segmentation~\cite{Girshick2014Segmentation,Hariharan2014Segmentation}, video classification~\cite{Karpathy2014Video}, and pose estimation~\cite{Toshev2014Pose}. CNN hierarchically builds up high-level semantic concepts from low-level visual features in a layer-by-layer manner based on convolution kernels which convolve pixels on local receptive fields. Among those tasks, object localization (or detection) is one of the fundamental problems in this research field. In object localization tasks, region-of-interests (ROIs), the most discriminative region in terms of semantic concepts, should be properly defined for each given image. A lot of training images with annotated bounding boxes or segmentation maps of ROIs are required in order to achieve good performance in object localization since those information gives strong prior in terms of exploring exact ROIs on test images~\cite{Sermanet2013Detection,Erhan2014Detection,Girshick2014Segmentation,Hariharan2014Segmentation}. However, a dataset with such location information is hard to obtain because it requires heavy annotation efforts. 
%Furthermore, in some cases, noisy annotation causes difficulties in training such a localization network. 

Weakly supervised learning for localization only uses a weak-labeled (i.e. image-level label) dataset which does not have any location information to localize objects in an image. In terms of finding common semantic features within a set of images having the same class label, this can be interpreted as a varient of multiple instance learning (MIL)~\cite{Andrews2002MIL,maron1998framework,foulds2010review}. 

Several previous works for CNN-based weakly supervised object localization have been presented with reasonable methods and good performances with the help of transfer learning from pre-trained networks~\cite{Oquab2015Weakly-GlobalMaxPool,Wu2015Weakly-Annotation,Pinheiro2015Weakly-LogSumExp}. Those approaches, however, require \textit{base networks} pre-trained on relatively well-localized datasets (e.g., ILSVRC classification dataset) which is able to extract discriminative regions appropriately from semantically similar datasets (e.g., VOC) while providing good initial seed for localization. In other words, they fine-tune good initial feature maps extracted from pre-trained networks with respect to the objectives of localization tasks. Weakly supervised localization methods which rely on those base networks cannot be used in the applications which do not have enough well-localized images. Medical image analysis is representative because it is impossible to obtain such pre-trained networks. Furthermore, it is not feasible to use base networks pre-trained on general images such as ILSVRC or VOC datasets since ROI characteristics of medical images are thoroughly different from general images.

In this work, we propose a self-transfer learning (STL) framework for fully weakly supervised localization. STL co-optimizes both classification and localization networks simultaneously in order to guide the localization network with the most discriminative features in terms of the classification task (see Figure~\ref{fig:overall_arch}). The term \textit{fully} means that the proposed method does not require not only the location information but also any types of pre-trained networks in a training stage, and the term \textit{self} stands for weight sharing between classification and localization networks.

Our contributions can be summarized as follows:
\begin{itemize}
\item We develop a fully weakly supervised learning framework based on CNN, a \textit{self-transfer learning}, which enables accurate ROI localization given only the image-level labeled dataset without any pre-trained model.
\item We show that a weakly supervised localization based on CNN without good initial weights is not effective by itself since errors are backpropagated through a restricted path or with insufficient information.
\item We conduct computational experiments on the medical application which is one of the most important areas in computer vision. We use chest X-rays and mammograms to show the localization performance of the proposed STL. It is shown that STL helps the localization network finding a good local optimum. 

\end{itemize}

The remainder of this paper is organized as follows. Section 2 presents previous works for the weakly supervised learning for object localization. In Section 3, the proposed STL framework is described in detail including its architecture and training scenarios. Section 4 shows experimental setup and results, and finally Section 5 concludes this paper.

%------------------------------------------------------------------------
\section{Related work}

There exist many studies to develop learning algorithms for object localization based on the weakly labeled dataset. Most previous works can be interpreted as the same framework, which use candidate regions extracted from an image and then select the correct localization among those regions~\cite{prev1,prev2,prev3,prev4}. In this work, recent methods based on CNN are considered since they have shown a promising performance on weakly supervised object localization~\cite{Oquab2015Weakly-GlobalMaxPool,Wu2015Weakly-Annotation,Pinheiro2015Weakly-LogSumExp}.

In a weakly supervised object localization task, we should find common features within a set of intra-class images, discriminate those intra-class features with each other, and define the most probable region in terms of target class. Transfer learning based on well pre-trained networks (preliminarily trained on different-but-similar datasets) helps to perform those challenging tasks, since pre-trained CNN properly defines ROIs based on discriminative convolutional filters already learned from semantically similar datasets~\cite{Oquab2015Weakly-GlobalMaxPool,Wu2015Weakly-Annotation,Pinheiro2015Weakly-LogSumExp}. 

In~\cite{Oquab2015Weakly-GlobalMaxPool}, convolutional feature extraction layers and adaptation layers are used for object localization. The convolutional feature extraction layers are pre-trained from the ImageNet dataset~\cite{Deng2009Imagenet}, so it appropriately extracts class-specific features from semantically similar datasets. From those feature maps, adaptation layers build up the class number of score maps. Adaptation layers consist of additional convolutional and global max pooling layers. The convolutional layer in the adaptation layers generates per-class score maps, and the most probable positions (with the highest activation) with respect to each class are pooled at each iteration in order to compute and backpropagate errors in a training phase. Since all the layers in this architecture are convolutional or pooling layers, rescaled input images can generate corresponding sizes of score maps. Each score map stands for the confidence level of existence of each object, i.e. per-class localization maps.

In~\cite{Wu2015Weakly-Annotation}, the authors train the deep multiple instance learning framework for jointly learning the object proposals and keywords simultaneously. They exploit parameters pre-trained on the ImageNet dataset~\cite{Deng2009Imagenet} for the object proposal network by assuming the object proposals per each image as a positive bag in the MIL framework. This assumption is quite reasonable, since the pre-trained network extracts class-specific feature representations properly. Given an image, the class probabilities for all object proposals are pooled at each iteration in order to be compared with the true label vector while propagating errors backward. In an inference stage, multiple object proposals with appropriate keywords enable the image-level auto-annotation. 

Inferring segmentation map is more challenging compared to object localization, since it should infer the class label per each pixel. In~\cite{Pinheiro2015Weakly-LogSumExp}, the authors propose the weakly supervised segmentation framework which uses datasets only with image-level labels in a training phase. A convolutional network pre-trained on the ImageNet dataset~\cite{Deng2009Imagenet} is used to extract feature maps from an input image. The class number of extracted feature maps (\textit{N}) are aggregated into a single \textit{N}-dimensional vector in the proposed aggregation layer to be compared with the true label. Compared to~\cite{Oquab2015Weakly-GlobalMaxPool}, the proposed aggregation layer adopts a \textit{Log-Sum-Exponential} pooling method which is a smooth version of the max pooling in order to fairly explore the entire feature maps. In an inference stage, the class number of feature maps extracted from an input image are used as initial maps for segmentation. By adding some segmentation priors (image-level prior and smoothing prior), they achieve good segmentation results compared to previous works related to weakly supervised object segmentation. 

Those weakly supervised approaches for object localization or segmentation are very helpful in general image domain because they do not require heavy annotation efforts as mentioned in the previous section. Strictly speaking, however, those are \textit{semi}-weakly instead of \textit{fully} weakly supervised framework, since they essentially require base networks pre-trained on different-but-similar datasets. Such networks pre-trained on well-localized datasets (e.g., ILSVRC classification dataset) can extract discriminative regions appropriately from different-but-similar datasets (e.g., VOC). Good initial feature maps extracted from pre-trained networks can be easily fine-tuned with respect to the objectives of localization tasks. Those previous works for \textit{semi}-weakly supervised localization cannot be used in the applications which do not have base networks pre-trained on semantically similar datasets.

%------------------------------------------------------------------------
\section{Self-transfer learning for weakly supervised learning}

In this section, we present our proposed STL framework for weakly supervised learning. STL consists of three main components: shared convolutional layers, fully connected layers (i.e. classifier), and localization layers (i.e. localizer) (see Figure~\ref{fig:overall_arch}). The key features of STL are twofold. First, it simultaneously propagates errors backward from both classifier and localizer to prevent the localizer from wandering a loss surface to find a local optimum. Second, an adjustable hyperparameter $\alpha$ is introduced to control the relative importance between classifier and localizer. 
%The overall systematic diagram of the STL framework is illustrated in Figure~\ref{fig:overall_arch}. 

\subsection{Weakly supervised localization based on CNN}

For the task of image classification, CNN works well by virtue of its ability to extract useful features which discriminate the classes. As we pointed out in Section 2, all previous works based on CNN utilize those features which already have good representation capability. 

The common strategy for weakly supervised localization based on CNN is to produce activation maps (in other words, score maps) for each class, and select or extract some representative activation value. Specifically, in case of a $K$-class classification problem, the network should give a $K$-dimensional class probability vector as an output to calculate errors using the true label vector. The intuitive way to make such output vector is to extract or select per-class representative activation from $K$-channel activation maps. The dimensions of those maps are automatically determined by a network architecture. For example, if the fully connected layers of \cite{Krizhevsky2012Classification} are replaced with convolutional layers and 512$\times$512 input image feeds into that network whose size of global receptive field is 224$\times$224, we can obtain $K\times$10$\times$10 activation maps since the global stride of the network is 32. If such a network is trained well, it is expected that a target object can be easily localized by examining the activation maps corresponding to its class.

To select or extract the representative activations for each class, typical pooling methods can be effectively used. In \cite{Oquab2015Weakly-GlobalMaxPool}, a global max pooling method is used and its classification and localization performances are verified in the domain of general images. Another choice can be a global average pooling method. It might be more effective if there are some classes which have no ROIs characterizing the image-level label. Those classes might be a background class in general image domain or a normal class in medical images considered in this work.
 
%Based on convolutional neural networks, multiple responses whose values are obtained from a single classifier which sees the subsampled image, called a global receptive field are generated. If that classifier has an ability to correctly classify objects which define class labels, we can easily localize the ROIs on the input image.

%To make a single channel activation map of each class based on feature activations, several approaches have been utilized. For instance, [] treated the fully connected layers as convolutions, which gives , and [] used additional convolution layers to obtain a score map for each class. Then, a single representative value is extracted from a score map by simple pooling methods such as max, average, etc. 

Those approaches can be interpreted as a variant of multiple instance learning (MIL), which is designed for classification where labels are associated with sets of instances, called bags, intead of individual instances. In image classification tasks, the full size image and its subsampled patches are considered as a bag and instances, respectively. For instance, if we use a global max pooling to select a representative value among activations of patches, it is equivalent to use a well-classified single patch for building the decision boundary. 
%Those approaches inherently suffer from seeing only subsampled region at each iteration. Therefore, it is hard to find good local optimum. In this paper, we propose the self-transfer learning framework which helps the localizer finding good features for both classification and localization. 

\subsection{Self-trasnfer learning}

The proposed STL framework is basically based on a joint learning of classifier and localizer. For successful training of localizer, the initial values of network weights (i.e. filters) are very important because the learned filters can guide the localizer. Without such good filters, it is hard to find a good local optimum since the localizer consistently concentrates on the subsampled region of original image whether it is a correct ROI or not. 

To overcome such a limitation, classifier and localizer are trained simultaneously based on a weighting strategy in our framework. Figure~\ref{fig:overall_arch} illustrates a systematic view of STL. In detail, it consists of shared convolutional layers \textbf{C}, classification layers \textbf{F}$_{\textbf{cls}}$, and localization layers \textbf{C}$_{\textbf{loc}}$. Two losses, \textbf{Loss}$_{\textbf{cls}}$ from classifier and \textbf{Loss}$_{\textbf{loc}}$ from localizer, are computed at the forward pass, and the weighted sum of those errors is propagated at the backward pass. The errors from classifier contribute to train the filters in an overall view, while those from localizer are backpropagated through the subsampled region which is the most important window to classify training set. At the early stage of training phase, the errors from classifier should be more weighted than those from localizer to prevent the localizer from falling in a bad local optimum. By reducing the effects of errors from localizer, good filters which have a discriminative power can be well trained although localizer fails to find objects associated with the class label. As training proceeds, the weight for localizer increases to focus on the subsampled region of input image. At this stage, the network's filters are fine-tuned for the task of localization.

\begin{figure}[t]
\begin{center}
\includegraphics[width=\columnwidth]{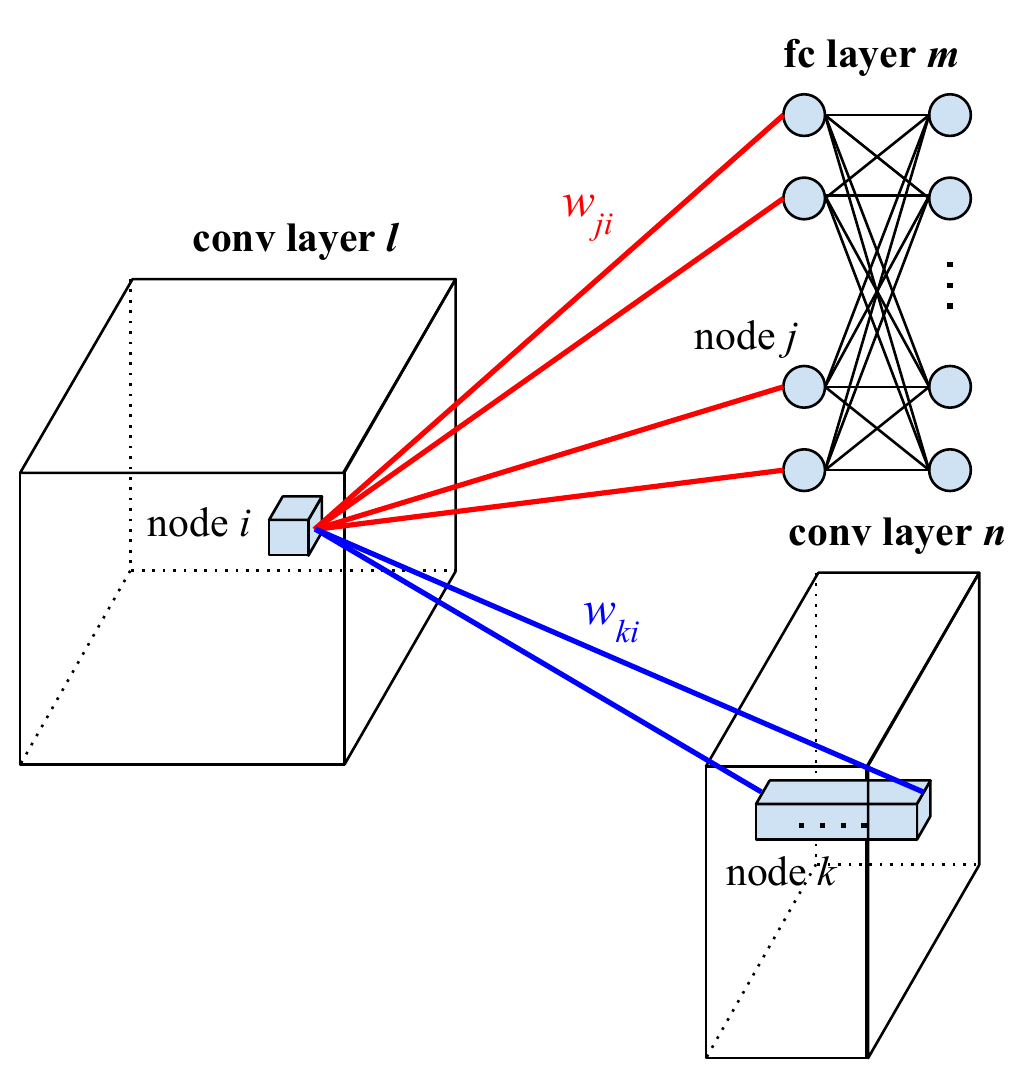}
%\includesvg{fig_backprop_tmp}
\end{center}
   \caption{Backpropagation at the end of shared convolutional layers. The weights directly related to compute the errors $\delta_i$ at the node $i$ are remained only for the visualization purpose.}
\label{fig:backprop}
\end{figure}

Consider a data set of $N$ input-target pairs $\{\mathbf{x}_i,\mathbf{t}_i\}_{i=1}^N$. $\mathbf{x}_i$ and $\mathbf{t}_i$ denote an $i$-th image and the corresponding $K$-dimensional true label vector, respectively, where $K$ represents the number of classes. Note that if $\mathbf{x}_i$ contains a single object, $\mathbf{t}_i$ will be an one-hot coded vector. Assuming an image with a single class label, our objective function to be optimized is a weighted sum of cross-entropy losses from classifier and localizer, which can be defined as follows:
\begin{equation}
\begin{split}
\label{eq:total_loss}
\text{\textbf{Loss}$_{\textbf{total}}$} &= (1-\alpha)\text{\textbf{Loss}$_{\textbf{cls}}$} + \alpha\text{\textbf{Loss}$_{\textbf{loc}}$}\\
&=-(1-\alpha) \sum_{i=1}^{N} {\mathbf{t}_{i}^\intercal\mathbf{log}({\mathbf{y}_i^{cls}}}) - \alpha \sum_{i=1}^{N} {\mathbf{t}_{i}^\intercal\mathbf{log}({\mathbf{y}_i^{loc}}})
\end{split}
\end{equation}
\noindent where $\mathbf{y}_i^{cls}$ and $\mathbf{y}_i^{loc}$ are $K$-dimensional class probability vectors from classifier and localizer, respectively, for $i	$-th input, and $\mathbf{log}(\cdot)$ denotes an element-wise log operation. Note that the loss function can be extended to be dealt with the case in which a single image has multiple labels~\cite{Oquab2015Weakly-GlobalMaxPool}.

The effect of the proposed STL can be explained by examining a backpropagation process at the end of shared convolutional layers \textbf{C}, which can be depicted as Figure~\ref{fig:backprop}. In this figure, the node $i$ represents a particular node in the convolutional layer $l$ which is connected with $H$ nodes in the fully connected layer $m$ and $K$ nodes in the convolutional layer $n$. Note that the layer $n$ is obtained by 1$\times$1 convolution on the layer $l$ as shown in Figure~\ref{fig:overall_arch} and $K$ is equal to the number of activation maps (i.e. the number of classes). If ReLU activation function is used for the node $i$, the backpropagated error $\delta_i$ at the node $i$ can be written as follows: 
\begin{equation}
\label{eq:delta}
\delta_i = \max (0, \delta^{cls}_i + \delta^{loc}_i)
\end{equation}
\noindent where
\begin{equation}
\label{eq:delta_fc}
\delta^{cls}_i =  \sum_{j=1}^{H} {w_{ji} \delta_j}
\end{equation}
\begin{equation}
\label{eq:delta_conv}
\delta^{loc}_i =  \sum_{k=1}^{K} {w_{ki} \delta_k}
\end{equation}
\noindent It should be noted that the relative importance between classifier and localizer is already reflected in the errors $\delta^{cls}_i$ and $\delta^{loc}_i$ through the weighted loss function defined as Equation~\ref{eq:total_loss}.

It can be seen that the errors $\delta^{loc}_i$ are backpropagated undesirably without $\delta^{cls}_i$ due to the special treatment, a global pooling, for activation maps in the layer $n$.
For instance, if a global max pooling is used to aggregate the activations within each activation map and the location corresponding to node $i$ in the layer $l$ is not selected as the maximum, all $\delta_k$'s to be backpropagated from the layer $n$ will be zero. Therefore, the computed errors of most of nodes in the layer $l$ except for the nodes whose locations correspond to the maximal responses for each activation map will be zero. In case of a global average pooling, zero errors will be merely replaced with a mean of errors. This situation is not certainly desirable, especially when we train the network from the bottom up (i.e. without pre-trained filters). By incorporating classifier into a network architecture, the shared convolutional layers \textbf{C} can be consistently improved even if the backpropagated errors $\delta^{loc}_i$ from localizer do not contribute to learn useful features.

The proposed STL framework is similar to well-known multi-task learning (MTL) \cite{vezhnevets2010towards,MTL2, MTL1}. MTL improves the performance of learning algorithms by learning classifiers for multiple related tasks jointly using the shared representation. We show that such MTL framework works well even if the network jointly learns exactly the same tasks. In this point of view, the proposed framework is more appropriate to be called as a multi-purpose learning since it has exactly the same tasks (i.e. classification with the same loss function), but have totally different purpose, one for classification and the other for localization.

\begin{table*}
\begin{center}
\def\arraystretch{1.2}
\small
\begin{tabular*}{\textwidth}{c|cccc|c|cccc|c}
\hline\hline
Testset & \multicolumn{5}{c|}{Shenzhen set} & \multicolumn{5}{c}{MC set} \\
\hline
Task & \multicolumn{4}{c|}{Classification} & Localization & \multicolumn{4}{c|}{Classification} & Localization \\
\hline
Metric & Accu. & AUC & AP(pos) & AP(neg) & AP & Accu. & AUC & AP(pos) & AP(neg) & AP \\
\hline
MaxPool          & 0.7855 &  0.8670 & 0.8920 & 0.8144 & 0.6982 & 0.7246 & 0.8047 & 0.8094 & 0.7971 & 0.6016\\
AvePool          & 0.7870 &  0.9069 & 0.9238 & 0.8879 & 0.6903 & 0.7029 & 0.7836 & 0.7522 & 0.8037 & 0.5119\\
STL+MaxPool & 0.8338 & 0.9166 &  0.9340 & 0.9004 & 0.7887 & 0.7681 & 0.8631 & 0.8484 & 0.8833 & 0.7956\\
STL+AvePool  & 0.8369 & 0.9267 & 0.9427 & 0.9064 & 0.8715 & 0.8406 & 0.8899 & 0.8838 & 0.8916 & 0.8070\\
\hline\hline
\end{tabular*}
\end{center}
\caption{Classification and localization performances for tuberculosis detection}
\label{table:tb_results}
\end{table*}

%\begin{table*}
%\begin{center}
%\def\arraystretch{1.2}
%\small
%\begin{tabular}{c|cccc|c}
%\hline\hline
%Testset & \multicolumn{5}{c}{MIAS set} \\
%\hline
%Task & \multicolumn{4}{c|}{Classification} & Localization \\
%\hline
%Metric & Accu. & AUC & AP(pos) & AP(neg) & AP  \\
%\hline
%MaxPool          & 0.5311 & 0.4862 & 0.3219 & 0.6526 & - \\
%AvePool          & 0.6615 & 0.6327 & 0.5439 & 0.7164 & 0.0952 \\
%STL+MaxPool & 0.6646 & 0.5356 & 0.4353 & 0.6644 & 0.1489 \\
%STL+AvePool  & 0.6957 & 0.6751 & 0.5753 & 0.7606 & 0.3256 \\
%\hline\hline
%\end{tabular}
%\end{center}
%\caption{Classification and localization performances for mammography}
%\label{table:mammo_results}
%\end{table*}

\begin{table}[t]
\begin{center}
\def\arraystretch{1.2}
\footnotesize
\begin{tabular*}{\columnwidth}{c|cccc|c}
\hline\hline
Testset & \multicolumn{5}{c}{MIAS set} \\
\hline
Task & \multicolumn{4}{c|}{Classification} & Loc. \\
\hline
Metric & Accu. & AUC & AP(pos) & AP(neg) & AP  \\
\hline
MaxPool          & 0.5311 & 0.4862 & 0.3219 & 0.6526 & - \\
AvePool          & 0.6615 & 0.6327 & 0.5439 & 0.7164 & 0.0952 \\
STL+MaxPool & 0.6646 & 0.5356 & 0.4353 & 0.6644 & 0.1489 \\
STL+AvePool  & 0.6957 & 0.6751 & 0.5753 & 0.7606 & 0.3256 \\
\hline\hline
\end{tabular*}
\end{center}
\caption{Classification and localization performances for mammography}
\label{table:mammo_results}
\end{table}

\section{Computational experiments}
In this section we use two medical image datasets, chest X-rays (CXRs) and mammograms, to evaluate the classification and localization performances of the proposed STL. CXRs and mammograms are very effective and frequently used for screening at the early stage of diagnosis process for tuberculosis and breast cancer. As mentioned in Section 1, such medical images generally do not have any additional information for localization (e.g., bounding boxes and/or segmentation maps for ROIs) except for image-level labels (e.g., normal, abnormal). However, it is very important not only to predict the precise image-level diagnosis result, but to provide finely localized ROIs for understanding of abnormalities. 

\begin{figure}[t]
\begin{center}
\begin{subfigure}{\columnwidth}
	\centering
	\includegraphics[width=0.9\columnwidth]{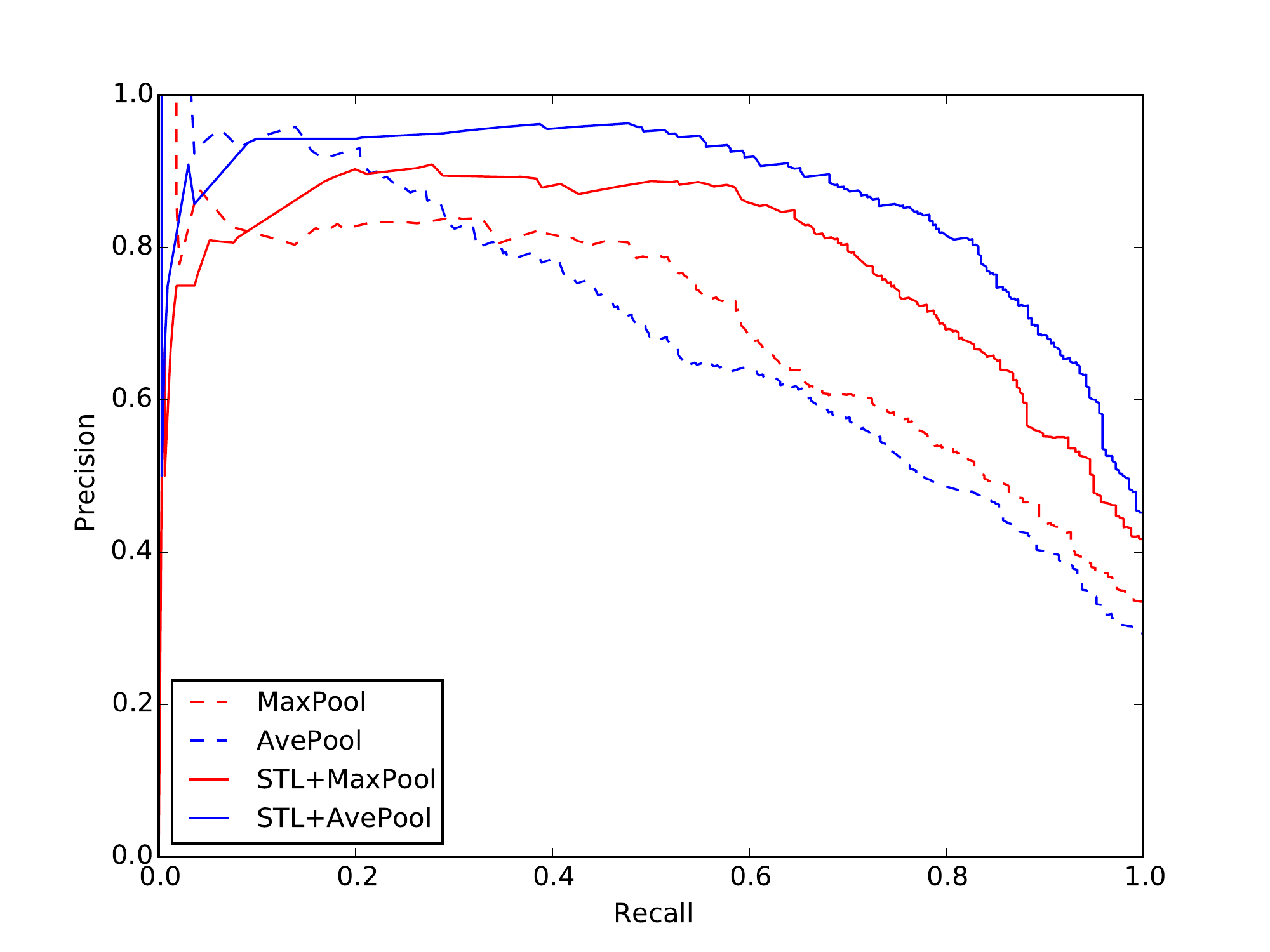}
	\caption{Shenzhen set}
\end{subfigure}
\begin{subfigure}{\columnwidth}
	\centering	
	\includegraphics[width=0.9\columnwidth]{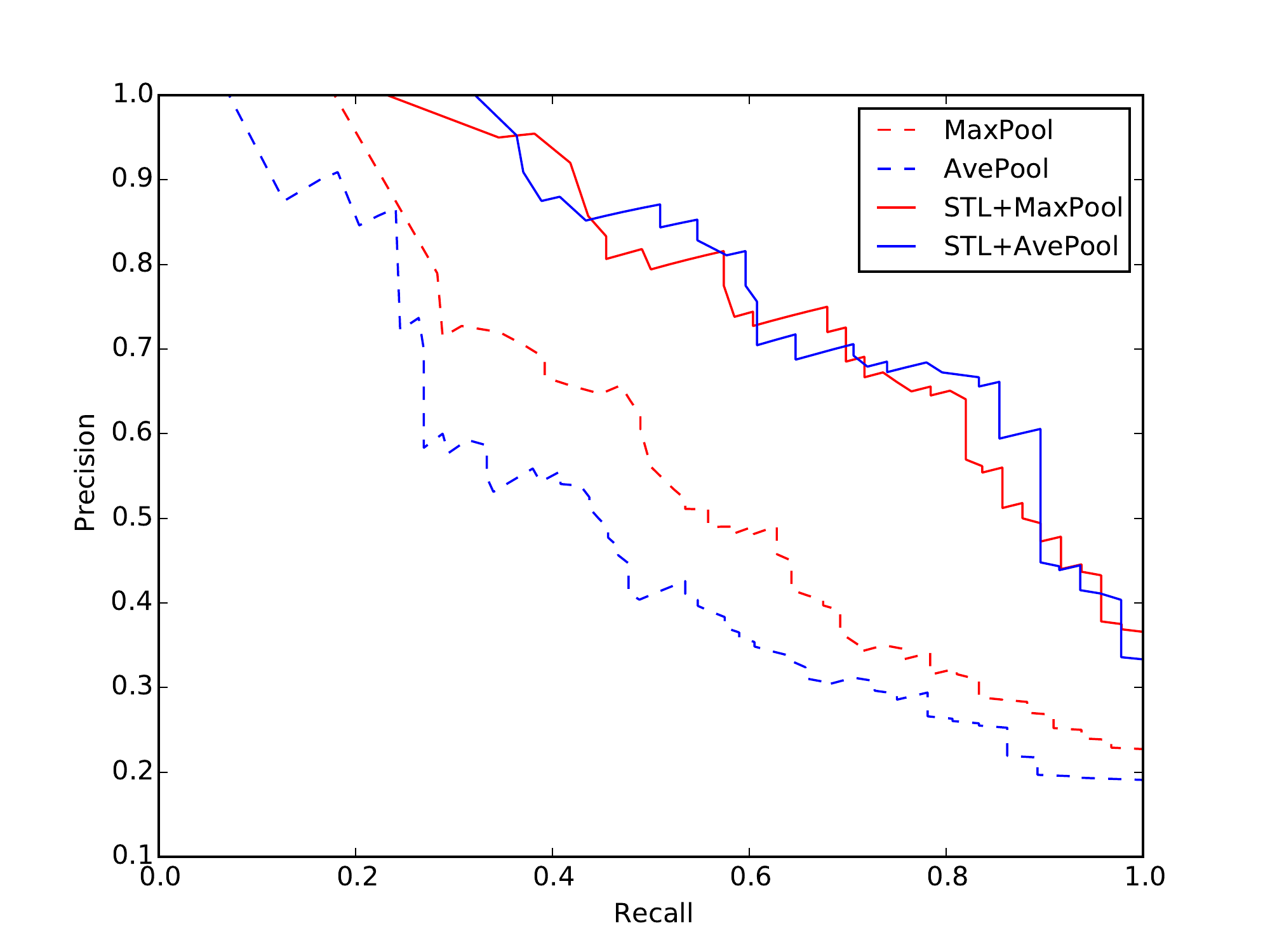}
	\caption{MC set}
\end{subfigure}
\begin{subfigure}{\columnwidth}
	\centering	
	\includegraphics[width=0.9\columnwidth]{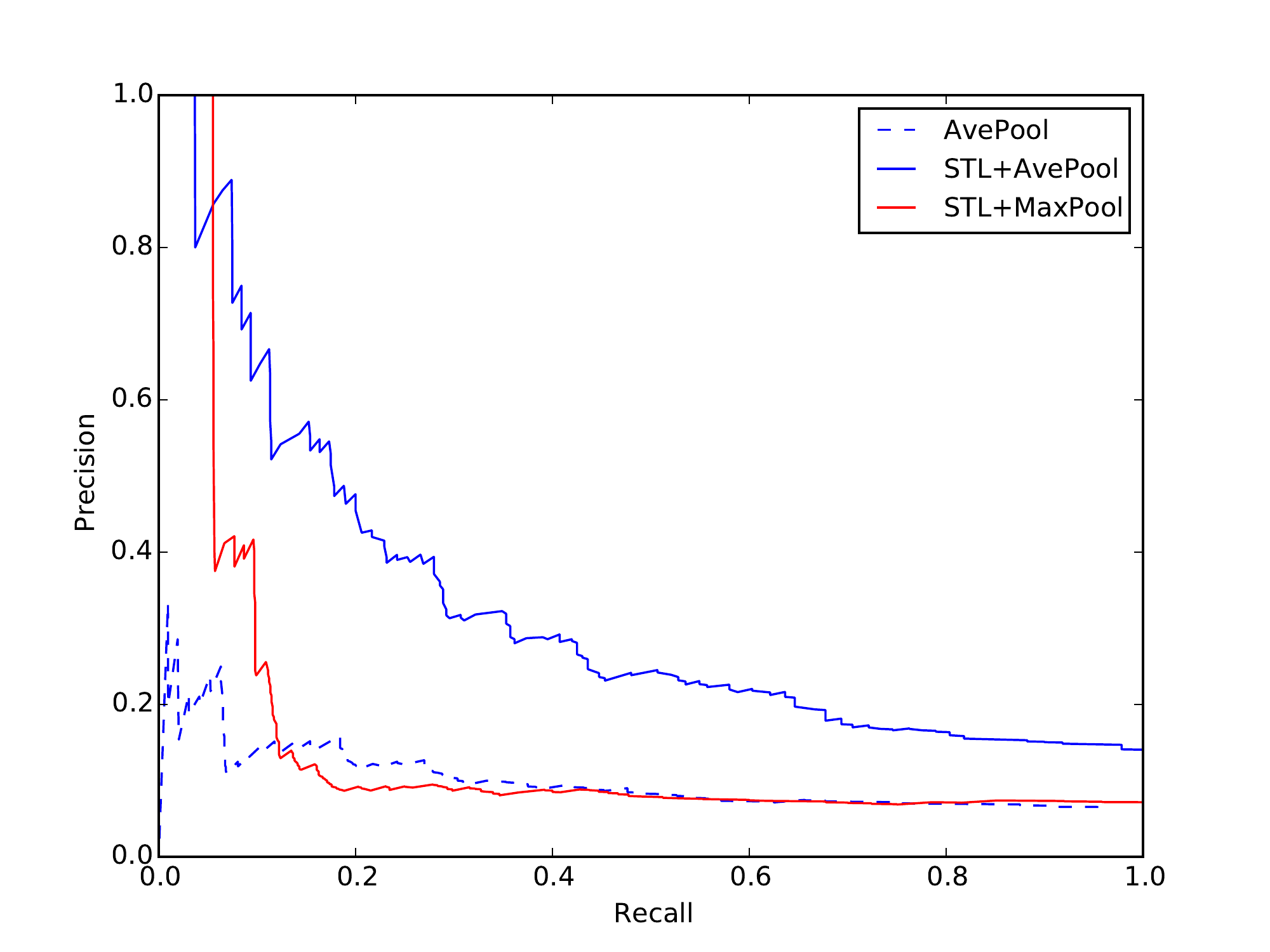}
	\caption{MIAS set}
\end{subfigure}
\end{center}
   \caption{Precision-recall curves for localization of positive class. Curve for MaxPool in (c) MIAS set is not shown due to no true positives regardless of a probability threshold.}
\label{fig:prcurve}
\end{figure}

\subsection{Experimental setup}
As abnormal ROIs usually have only a small portion of the entire image, the network should be trained by high resolution images enough to capture the ROIs into its global receptive field. Therefore, all training CXRs and mammograms are resized to 500$\times$500. The network architecture used in this experiment is slightly modified based on the network from \cite{Krizhevsky2012Classification}. We add one convolutional layer (i.e. the 6th convolutional layer) since the resolution of the input image is relatively high compared to images for general object classification tasks. Also, we set the number of hidden nodes in the fully connected layers to 2048. For localizer, 1$\times$1 convolution operation performs on the added convolutional layer, and therefore 15$\times$15 activation maps for each class are obtained. The weights in each layer are initialized from a zero-mean Gaussian distribution with standard deviation 0.01, and initial biases are set to 0.

In this experiment, the number of activation maps is two since we are dealing with classification of two classes, normal and abnormal. To verify the effectiveness of STL, two typical pooling methods, max and average poolings, are applied globally to the activation maps, and their performance improvements with STL are examined. As depicted in Figure~\ref{fig:overall_arch}, the \textit{softmax} loss function is used for both classifier and localizer. 

All the experiments in this work are performed using Caffe~\cite{Jia2014Caffe}. Each training set is randomly divided into 80\% for training and 20\% for validation, and the final model with the best validation accuracy is selected for performance evaluation. We consider an initial learning rate $\lambda=0.01$ and it is decreased by a factor of 2 for every 30 epochs. The network is trained via stochastic gradient descent with momentum 0.9 and the minibatch size is set to 64. The weight decay parameter is determined by a grid search through the comparison of validation accuracy. There is an additional hyperparameter $\alpha$ on STL to determine the level of importance between classifier and localizer. We set its initial value to 0.1 so that the network more focuses on learning the representative features at the early stage, and it is increased to 0.9 after 60 epochs to fine-tune the localizer.

\subsection{Performance measures}
To compare the classification performance, an area under characteristic curve (AUC), accuracy and average precision (AP) of each class are used. For STL, class probabilities obtained from localizer is used for measuring performance. For a localization task, a similar performance metric in \cite{Oquab2015Weakly-GlobalMaxPool} is used. It is based on AP, but the difference is the way to count true positives and false positives. In classification, it is a true positive if its class probability exceeds some threshold. To measure the localization performance under this metric, the test image whose class probability is greater than some threshold (i.e. a true positive in case of classification) but the maximal response in the activation map does not fall within the ground truth annotations allowing some tolerance is counted as a false positive.

In our experiment, only positive class is considered for localization AP since there is no ROI on negative class. First, the activation map of positive class is resized to the size of original image via simple bilinear interpolation, then it is examined whether the maximal response falls into the ground truth annotations within 16 pixels tolerances which is a half of the global stride 32 of the considered network architecture. If the response is located inside true annotations, the test image is counted as a true positive. If not, it is counted as a false positive.

\subsection{Tuberculosis detection}

Tuberculosis (TB) is one of the major global health threats. Many curable TB patients in the developing countries are obliged to die because of delayed diagnosis, partly by the lack of radiography and radiologists. Therefore, developing a computer-aided diagnosis (CAD) system for TB screening can contribute to early diagnosis of TB, which results in prevention of deaths from TB. 

We use three CXRs datasets, namely KIT, Shenzhen and MC sets in this experiment. All the CXRs used in this work are de-identified by the corresponding image providers. KIT set contains 10,848 DICOM images, consisting of 7,020 normal and 3,828 abnormal (TB) cases, from the Korean Institute of Tuberculosis (KIT) under Korean National Tuberculosis Association (KNTA), South Korea. Shenzhen and MC sets are available limited to research purpose \cite{Candemir2014TB,Jaeger2013TB,Jaeger2014TB}. Shenzhen set has 326 normal and 336 TB cases from Shenzhen No. 3 People's Hospital, Guangdong Medical College, Shenzhen, China. Finally, MC set was collected from National Library of Medicine, National Institutes of Health, Bethesda, MD, USA. It consists of 80 normal and 58 TB cases.

We train the models using the KIT set, and test the classification and localization performances using the Shenzhen and MC sets. To evaluate the localization performance, we obtain their detailed annotations from the TB clinician since the testsets, Shenzhen and MC sets, do not contain any annotations for TB ROIs. 
Table~\ref{table:tb_results} summarizes the experimental results. For both classification and localization tasks, STL consistently outperforms other methods. The best performance model is STL+AvePool. A global average pooling works well in this experiment since images of negative class act as a background. 
Since the value of localization AP is always less than that of classification AP (refer to the definition of localization AP in Section 4.2), it is important to see the improvement ratio. For a global average pooling, the localization APs are improved about 26\% and 58\% for Shenzhen and MC sets, respectively, while the improvement of classification APs for positive class are about  2\% and 17\%. This means that STL certainly assists localizer to find the most important ROIs which define the class label. 
Precision-recall curve is shown in Figure~\ref{fig:prcurve}. The left half of Figure~\ref{fig:examples} shows the representative examples among the test sets.

\subsection{Mammography}

We use two public mammography databases, called Digital Database for Screening Mammography (DDSM)~\cite{heath1998DDSM,heath2000DDSM} and Mammographic Image Analysis Society (MIAS)~\cite{Suckling1994Mammo}, in this experiment. The DDSM and MIAS are used for training and testing, respectively. 
We preprocess DDSM images to have two labels, positive (abnormal) and negative (normal). Originally, abnormal mammographic images contain several types of abnormalities such as masses, microcalcification, etc. We merge all types of abnormalities into positive class to distinguish any abnormalities from normal, thus the number of positive and negative images are 4025 and 6338 respectively in the training set (DDSM). In test set (MIAS), there are 112 positive and 210 negative images. Note that we do not use any additional information except for image-level labels for training networks although the training set (DDSM) has boundary information for abnormal ROIs. The boundary information of test set (MIAS) is utilized to evaluate the localization performance. 

\begin{figure*}
\begin{center}
\includegraphics[width=\linewidth]{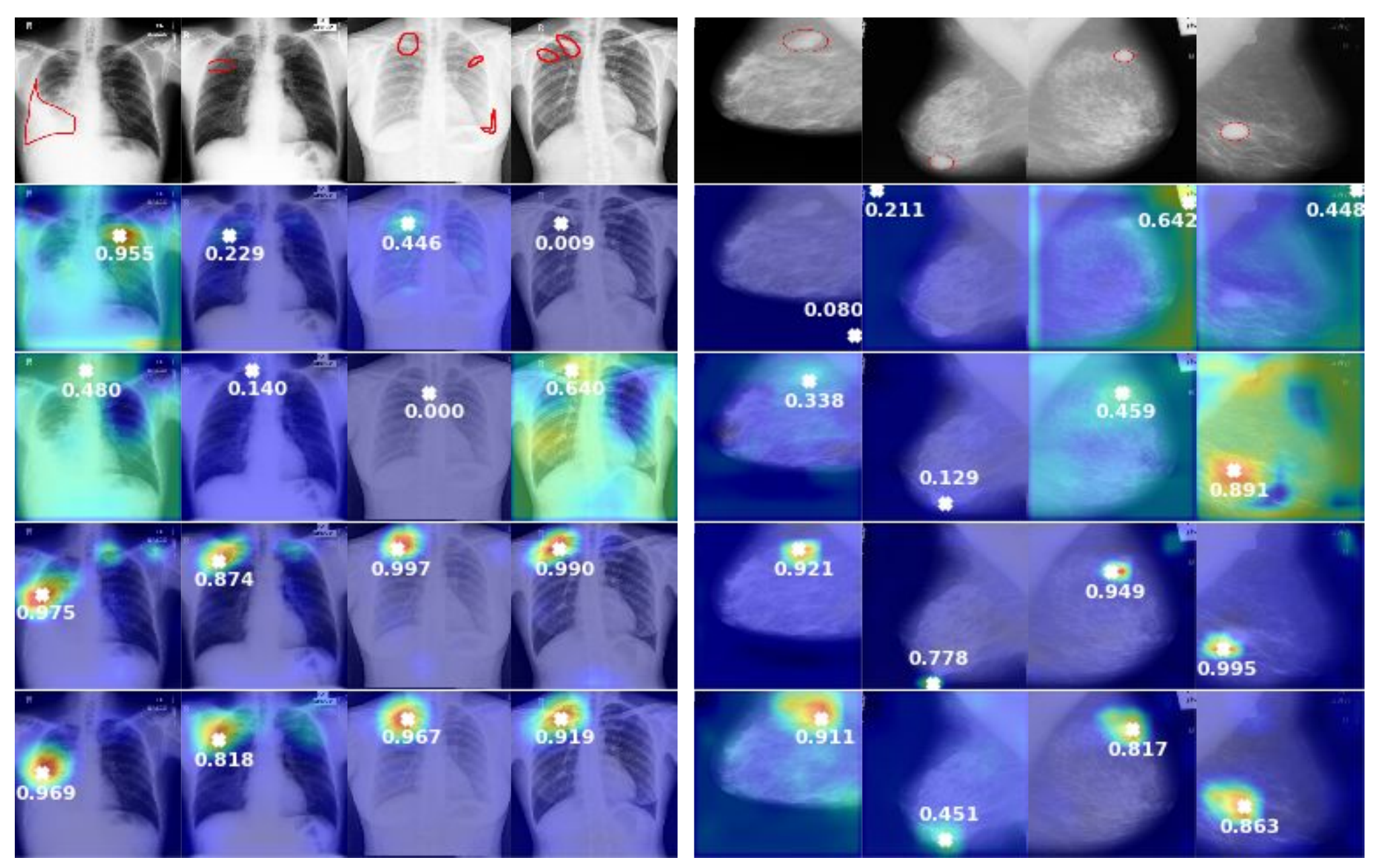}
\end{center}
   \caption{Localization examples for chest X-rays and mammograms. Top row shows test images with groud-truth annotations. The belows represent the results from MaxPool, AvePool, STL+MaxPool and STL+AvePool in a sequential order. The activation map for positive class is linearly scaled to the range between 0 and the maximum probability.}
\label{fig:examples}
\end{figure*}

Table~\ref{table:mammo_results} reports the classification and localization results. As we can see, classification of mammograms is much difficult compared to TB detection. First of all, mammograms used for training are low quality images which contain some degree of artifact and distortion generated from the scanning process for creating digital images from films. Moreover, this task is inherently complicated since there also exist quite a few irregular patterns in normal class caused by various shapes and characteristics of fatty tissues. Nevertheless, it is confirmed that STL is significantly better than other methods for both classification and localization. 
Again, for a global average pooling, the localization performance is improved about 242\% while the classification performance is improved about 6\%. For a global max pooling without STL, training loss is not decreased at all, i.e., it cannot be trained. Therefore, the localization performance of that is not reported in Table~\ref{table:mammo_results} and Figure~\ref{fig:prcurve} since there are no true positives for all probability thresholds.
Figure~\ref{fig:prcurve} and Figure~\ref{fig:examples} show precision-recall curve and the representative examples among the test sets.

\section{Conclusion}

In this work, we propose a novel framework STL which enables training CNN for object localization without neither any location information nor pre-trained models. Our framework jointly learns both classifier and localizer using a weighted loss as an objective function for the purpose of preventing localizer from falling in a bad local optimum. Self-transfer is realized via a weight controlling the relative importance between classifier and localizer. Also, the effect of classifier on localizer is discussed to provide the rationale behind the advantages of the proposed framework. Computational experiments for medical vision tasks given only image-level labels show that the proposed framework outperforms the existing approaches in terms of both classification and localization performance metrics. 

\newpage

{\small
\bibliographystyle{ieee}
\bibliography{egbib}

\begin{thebibliography}{10}\itemsep=-1pt

\bibitem{Andrews2002MIL}
S.~Andrews, I.~Tsochantaridis, and T.~Hofmann.
\newblock Support vector machines for multiple-instance learning.
\newblock In {\em Advances in Neural Information Processing Systems (NIPS)},
  pages 561--568, 2002.

\bibitem{Candemir2014TB}
S.~Candemir, S.~Jaeger, K.~Palaniappan, J.~P. Musco, R.~K. Singh, Z.~Xue,
  A.~Karargyris, S.~Antani, G.~Thoma, and C.~J. McDonald.
\newblock Lung segmentation in chest radiographs using anatomical atlases with
  nonrigid registration.
\newblock {\em IEEE Transactions on Medical Imaging}, 33(2):577--590, 2014.

\bibitem{Deng2009Imagenet}
J.~Deng, W.~Dong, R.~Socher, L.-J. Li, K.~Li, and L.~Fei-Fei.
\newblock Imagenet: A large-scale hierarchical image database.
\newblock In {\em IEEE Conference on Computer Vision and Pattern Recognition
  (CVPR)}, pages 248--255, 2009.

\bibitem{Erhan2014Detection}
D.~Erhan, C.~Szegedy, A.~Toshev, and D.~Anguelov.
\newblock Scalable object detection using deep neural networks.
\newblock In {\em IEEE Conference on Computer Vision and Pattern Recognition
  (CVPR)}, pages 2155--2162, 2014.

\bibitem{foulds2010review}
J.~Foulds and E.~Frank.
\newblock A review of multi-instance learning assumptions.
\newblock {\em The Knowledge Engineering Review}, 25(01):1--25, 2010.

\bibitem{Girshick2014Segmentation}
R.~Girshick, J.~Donahue, T.~Darrell, and J.~Malik.
\newblock Rich feature hierarchies for accurate object detection and semantic
  segmentation.
\newblock In {\em IEEE Conference on Computer Vision and Pattern Recognition
  (CVPR)}, pages 580--587, 2014.

\bibitem{Hariharan2014Segmentation}
B.~Hariharan, P.~Arbel{\'a}ez, R.~Girshick, and J.~Malik.
\newblock Simultaneous detection and segmentation.
\newblock In {\em European Conference on Computer Vision (ECCV)}, pages
  297--312, 2014.

\bibitem{heath1998DDSM}
M.~Heath, K.~Bowyer, D.~Kopans, P.~Kegelmeyer~Jr, R.~Moore, K.~Chang, and
  S.~Munishkumaran.
\newblock Current status of the digital database for screening mammography.
\newblock In {\em Proceedings of the Fourth International Workshop on Digital
  Mammography}, pages 457--460, 1998.

\bibitem{heath2000DDSM}
M.~Heath, K.~Bowyer, D.~Kopans, R.~Moore, and W.~P. Kegelmeyer.
\newblock The digital database for screening mammography.
\newblock In {\em Proceedings of the 5th international workshop on digital
  mammography}, pages 212--218, 2000.

\bibitem{Jaeger2014TB}
S.~Jaeger, A.~Karargyris, S.~Candemir, L.~Folio, J.~Siegelman, F.~Callaghan,
  Z.~Xue, K.~Palaniappan, R.~K. Singh, S.~Antani, et~al.
\newblock Automatic tuberculosis screening using chest radiographs.
\newblock {\em IEEE Transactions on Medical Imaging}, 33(2):233--245, 2014.

\bibitem{Jaeger2013TB}
S.~Jaeger, A.~Karargyris, S.~Candemir, J.~Siegelman, L.~Folio, S.~Antani, and
  G.~Thoma.
\newblock Automatic screening for tuberculosis in chest radiographs: a survey.
\newblock {\em Quantitative Imaging in Medicine and Surgery}, 3(2):89--99,
  2013.

\bibitem{Jia2014Caffe}
Y.~Jia, E.~Shelhamer, J.~Donahue, S.~Karayev, J.~Long, R.~Girshick,
  S.~Guadarrama, and T.~Darrell.
\newblock Caffe: Convolutional architecture for fast feature embedding.
\newblock In {\em Proceedings of ACM International Conference on Multimedia},
  pages 675--678, 2014.

\bibitem{Karpathy2014Video}
A.~Karpathy, G.~Toderici, S.~Shetty, T.~Leung, R.~Sukthankar, and L.~Fei-Fei.
\newblock Large-scale video classification with convolutional neural networks.
\newblock In {\em IEEE Conference on Computer Vision and Pattern Recognition
  (CVPR)}, pages 1725--1732, 2014.

\bibitem{Krizhevsky2012Classification}
A.~Krizhevsky, I.~Sutskever, and G.~E. Hinton.
\newblock Imagenet classification with deep convolutional neural networks.
\newblock In {\em Advances in Neural Information Processing Systems (NIPS)},
  pages 1097--1105, 2012.

\bibitem{maron1998framework}
O.~Maron and T.~Lozano-P{\'e}rez.
\newblock A framework for multiple-instance learning.
\newblock In {\em Advances in Neural Information Processing Systems (NIPS)},
  pages 570--576, 1998.

\bibitem{Oquab2015Weakly-GlobalMaxPool}
M.~Oquab, L.~Bottou, I.~Laptev, and J.~Sivic.
\newblock Is object localization for free?--weakly-supervised learning with
  convolutional neural networks.
\newblock In {\em IEEE Conference on Computer Vision and Pattern Recognition
  (CVPR)}, pages 685--694, 2015.

\bibitem{prev2}
M.~Pandey and S.~Lazebnik.
\newblock Scene recognition and weakly supervised object localization with
  deformable part-based models.
\newblock In {\em IEEE International Conference on Computer Vision (ICCV)},
  pages 1307--1314, 2011.

\bibitem{Pinheiro2015Weakly-LogSumExp}
P.~O. Pinheiro and R.~Collobert.
\newblock From image-level to pixel-level labeling with convolutional networks.
\newblock In {\em IEEE Conference on Computer Vision and Pattern Recognition
  (CVPR)}, pages 1713--1721, 2015.

\bibitem{Sermanet2013Detection}
P.~Sermanet, D.~Eigen, X.~Zhang, M.~Mathieu, R.~Fergus, and Y.~LeCun.
\newblock Overfeat: Integrated recognition, localization and detection using
  convolutional networks.
\newblock {\em arXiv preprint arXiv:1312.6229}, 2013.

\bibitem{prev4}
S.~Singh, A.~Gupta, and A.~Efros.
\newblock Unsupervised discovery of mid-level discriminative patches.
\newblock In {\em European Conference on Computer Vision (ECCV)}, pages 73--86,
  2012.

\bibitem{Suckling1994Mammo}
J.~Suckling, J.~Parker, D.~Dance, S.~Astley, I.~Hutt, C.~Boggis, I.~Ricketts,
  E.~Stamatakis, N.~Cerneaz, S.~Kok, et~al.
\newblock The mammographic image analysis society digital mammogram database.
\newblock In {\em Exerpta Medica. International Congress Series}, volume 1069,
  pages 375--378, 1994.

\bibitem{Szegedy2015Classification}
C.~Szegedy, W.~Liu, Y.~Jia, P.~Sermanet, S.~Reed, D.~Anguelov, D.~Erhan,
  V.~Vanhoucke, and A.~Rabinovich.
\newblock Going deeper with convolutions.
\newblock In {\em IEEE Conference on Computer Vision and Pattern Recognition
  (CVPR)}, pages 1--9, 2015.

\bibitem{Toshev2014Pose}
A.~Toshev and C.~Szegedy.
\newblock Deeppose: Human pose estimation via deep neural networks.
\newblock In {\em IEEE Conference on Computer Vision and Pattern Recognition
  (CVPR)}, pages 1653--1660, 2014.

\bibitem{vezhnevets2010towards}
A.~Vezhnevets and J.~M. Buhmann.
\newblock Towards weakly supervised semantic segmentation by means of multiple
  instance and multitask learning.
\newblock In {\em IEEE Conference on Computer Vision and Pattern Recognition
  (CVPR)}, pages 3249--3256, 2010.

\bibitem{prev3}
S.~Vijayanarasimhan and K.~Grauman.
\newblock Keywords to visual categories: Multiple-instance learning forweakly
  supervised object categorization.
\newblock In {\em IEEE Conference on Computer Vision and Pattern Recognition
  (CVPR)}, pages 1--8, 2008.

\bibitem{prev1}
C.~Wang, W.~Ren, K.~Huang, and T.~Tan.
\newblock Weakly supervised object localization with latent category learning.
\newblock In {\em European Conference on Computer Vision (ECCV)}, pages
  431--445, 2014.

\bibitem{MTL2}
X.~Wang, C.~Zhang, and Z.~Zhang.
\newblock Boosted multi-task learning for face verification with applications
  to web image and video search.
\newblock In {\em IEEE Conference on Computer Vision and Pattern Recognition
  (CVPR)}, pages 142--149, 2009.

\bibitem{Wu2015Weakly-Annotation}
J.~Wu, Y.~Yu, C.~Huang, and K.~Yu.
\newblock Deep multiple instance learning for image classification and
  auto-annotation.
\newblock In {\em IEEE Conference on Computer Vision and Pattern Recognition
  (CVPR)}, pages 3460--3469, 2015.

\bibitem{MTL1}
T.~Zhang, B.~Ghanem, S.~Liu, and N.~Ahuja.
\newblock Robust visual tracking via multi-task sparse learning.
\newblock In {\em IEEE Conference on Computer Vision and Pattern Recognition
  (CVPR)}, pages 2042--2049, 2012.

\end{thebibliography}
}

\end{document}